\def\year{2020}\relax
\documentclass[letterpaper]{article} 
\usepackage{aaai20}  
\usepackage{times}  
\usepackage{helvet} 
\usepackage{courier}  
\usepackage[hyphens]{url}  
\usepackage{graphicx} 
\usepackage{graphicx}  
\usepackage{amssymb}
\usepackage [usenames, dvipsnames] {color}
\usepackage{subfig}
\pdfoutput=1
\title{SPSTracker: Sub-Peak Suppression of Response Map for Robust Object Tracking}

\author{Qintao Hu\textsuperscript{\rm 1,2,3}\thanks{Both authors contributed equally.}, Lijun Zhou\textsuperscript{\rm 2,3,4}\footnotemark[1], Xiaoxiao Wang\textsuperscript{\rm 5}, Yao Mao\textsuperscript{\rm 1,2}\thanks{Corresponding author.}, Jianlin Zhang\textsuperscript{\rm 1,2}, Qixiang Ye\textsuperscript{\rm 3}\footnotemark[2]\\ 
\textsuperscript{\rm 1}Key Laboratory of Optical Engineering, Chinese Academy of Sciences, Chengdu , China\\ 
\textsuperscript{\rm 2} Institute of Optics and Electronics, Chinese Academy of Sciences, Chengdu , China \\ 
\textsuperscript{\rm 3}University of Chinese Academy of Sciences, Beijing , China\\ 
\textsuperscript{\rm 4}TU Kaiserslautern, Kaiserslautern, Germany\\
\textsuperscript{\rm 5}University of California, Davis, USA\\
\{huqintao16, zhoulijun16\}@mails.ucas.edu.cn, xxwa@ucdavis.edu,  \{maoyao,jlin\}@ioe.ac.cn, qxye@ucas.ac.cn
}

\begin{document}

\maketitle

\begin{abstract}

Modern visual trackers usually construct online learning models under the assumption that the feature response has a Gaussian distribution with target-centered peak response. Nevertheless, such an assumption is implausible when there is progressive interference from other targets and/or background noise, which produce sub-peaks on the tracking response map and cause model drift. In this paper, we propose a rectified online learning approach for sub-peak response suppression and peak response enforcement and target at handling progressive interference in a systematic way. Our approach, referred to as SPSTracker, applies simple-yet-efficient Peak Response Pooling (PRP) to aggregate and align discriminative features, as well as leveraging a Boundary Response Truncation (BRT) to reduce the variance of feature response. By fusing with multi-scale features, SPSTracker aggregates the response distribution of multiple sub-peaks to a single maximum peak, which enforces the discriminative capability of features for robust object tracking. Experiments on the OTB, NFS and VOT2018 benchmarks demonstrate that SPSTrack outperforms the state-of-the-art real-time trackers with significant margins\footnote{The code is available at. {\color{Magenta}github.com/TrackerLB/SPSTracker}}
\end{abstract}

\begin{figure}[htb]
\begin{center}
{\includegraphics[width=.95\columnwidth]{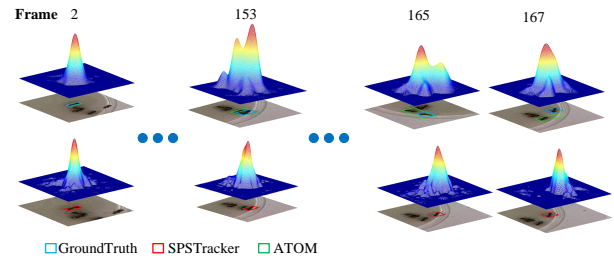}}
\end{center}
\caption{Comparison of our approach with the state-of-the-art ATOM tracker~\cite{ATOM}. For the interference from multiple targets, ATOM (first row) produces response maps of multiple sub-peaks (frame 153 to frame 165), which cause model drift. In contrast, our approach (second row) aggregates the response distribution of multiple sub-peaks to a single maximum peak, which leads to robust target tracking. (Best viewed in color and with zoom in)
}
\label{fig:motivation}
\end{figure}

\section{Introduction}

In the past few years, deep convolutional neural networks (CNNs) have significantly improved the performance of visual object tracking, by providing frameworks for end-to-end representation learning~\cite{HCF,Danelljan2016}, online correlation filter learning~\cite{KCF2015,CCOT}, and discriminative classifier learning ~\cite{ECO,Qi2016,Nam2015,Nam2016,Han}. 
However, CNN-based trackers suffer the performance degradation caused by the multi-target occlusion, appearance variance and/or background noise. Especially during the tracking procedure, the neighboring targets and background noise could introduce progressive interference and result in the vital model drift, as shown in Fig.~\ref{fig:motivation}(up), particularly when objects have scale variations and complex motions.

To mitigate the interference, the Siamese network structure was introduced to improve the discriminative capacity of trackers by extensively training the network~\cite{Tao2016,Xu2017}. 
EArly-Stopping Tracker incorporates object representation decision-making policies with the reinforcement learning method~\cite{Huang2017}. Nevertheless, these approaches usually rely on using additional data for offline training and lack the capability to adapt trackers adaptability in complex conditions. 

To conquer the issue, the dynamic Siamese network~\cite{Guo2017} uses a fast transformation learning method to model target appearance variation and handles background suppression from previous frames. The ATOM tracker~\cite{ATOM} combines offline pre-training with online learning in a multi-task learning framework by incorporating the objectives of target localization and target-background classification. 
While incorporating high-level knowledge into the target estimation through extensive offline learning, these methods remain overlooking the progressive interference from context area in a systematic manner. How to directly model the interference and regularize the tracking response distribution is still a open problem.

In this paper, we propose a simple-yet-effective approach, referred to as SPSTracker for robust object tracking. Our motivation is based on the observation that most failure tracking is caused by the interference around the target. Such interference produces multi-peak tracking response, and the sub-peak may progressively ``grow" and eventually cause model drift. Therefore ,we propose suppressing the sub-peaks to aggregating a single-peak response, with the aim of preventing model drift from the perspective of tracking response regularization.

\begin{figure*}
\begin{center}
\includegraphics[width= 1.6\columnwidth]{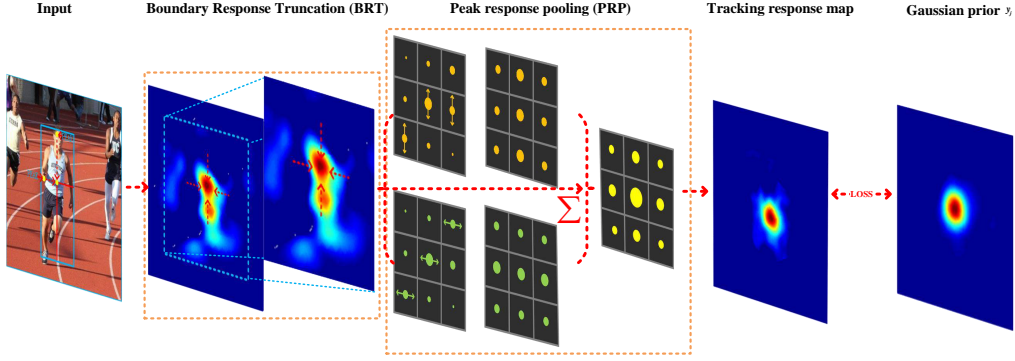}
\end{center}
  \caption{Illustration of Boundary Response Truncation (BRT) and  Peak Response Pooling (PRP) modules. First, with BRT, we clips the feature response map meanwhile aggregates the single-peak response. Then ,with PRP, we sum the horizontal and vertical pooling maps to aggregate multiple sub-peaks(the surrounding small dots in pooling maps) into a single centered peak(large dot) for target tracking. (Best viewed in color)}
\label{fig:method}
\end{figure*}

Specifically, we introduce a Peak Response Pooling (PRP) module, which concentrates the maximum values of tracking response into the geometric centers of targets, as shown in Fig.~\ref{fig:method}. The pooling procedure is implemented by an efficient maximization and substitution operation on the tracking response maps. During the network forward procedure, PRP aggregates multiple sub-peaks into a single centered peak for target tracking. During the backward propagation procedure, the response map with a single peak guides the online learning (fine-tuning) to explore discriminative features. 

Based on PRP, we further propose the Boundary Response Truncation (BRT) operation to clip the response map by simply setting the values of the pixels far away from the peak response to be zero. The operation reduces the variance of feature response map meanwhile further aggregates the single-peak response. If the response map is approximated as a Gaussian distribution, PRP targets at aggregating the mean values while BRT reducing the variance. RPR together with BRT facilitates the model to learn an enforced response map for robust object tracking.

SPSTracker is built upon the CNN framework with a target classification branch and a target localization branch atop the convolutional layers.
The classification network, equipped with the PRP and BRT modules, identifies the coarse locations (bounding boxes). These coarse locations are further fed to the target localization branch to estimate the precise target location.

The main contributions of this work can be summarized as follows:
\begin{itemize}
    \item An Sub-Peak Suppression tracker (SPSTracker) is presented to reduce the risk of model drift by online suppressing the potential sub-peaks while aggregating the maximum peak response.
    \item A simple-yet-efficient Peak Response Pooling (PRP) module is proposed to aggregate and align discriminative features, and a Boundary Response Truncation (BRT) module is designed to reduce the variance of feature response.
    \item Our proposed tracker achieves leading performance on six benchmarks, including OTB2013, OTB2015, OTB50, VOT2016, VOT2018 and NFS. In particular, we improve the state-of-the-arts on the VOT2016 and VOT2018 benchmarks with significant margins.
    
\end{itemize}


\section{Related Work}

The research about visual object tracking has a long history. Modern object trackers were usually constructed on three kinds of methods, including correlation filtering, online classification, and metric learning. With the rise of deep neural networks, these methods have been integrated with feature learning in an end-to-end manner.

\textbf{Correlation Filters.} The filtering procedure refers to matching templates with the Gaussian distribution to track targets of various appearance variation. The key to their success is the ability to efficiently exploit available negative data by including all shifted versions of a training sample. By introducing CNNs, the representative capacity of correlation filters has been greatly improved. DeepSRDCF~\cite{SRDCF} fed the features from the pre-trained CNN to a correlation filter and introduced spatial regularization on the basis of KCF~\cite{KCF}, mitigating the boundary effect. CCOT~\cite{CCOT} and ECO~\cite{ECO} proposed the implicit interpolation model to pose the learning problem in the continuous spatial domain, leading to efficient integration of multi-resolution deep features.

\textbf{Online Classification.} Tracking can also be formulated as an online classification problem. DeepTrack~\cite{DeepTrack} leveraged a sample selection mechanism and a lazy updating scheme to learn online classifiers. FCNT~\cite{FCNT} utilized hierarchical convolutional features to construct a network which handles various interference. CNN-SVM~\cite{CNN-SVM} used the pre-trained deep convolutional network to extract features of the target, and then used SVM to perform online target-background classification. These methods fully utilized the representation capabilities of deep learning features and the discriminative capacity of online classifiers. However, they often overlook the problem of accurate target state estimation.

\textbf{Metric Learning.}  To facilitate state estimation, the tracking problem was formulated in the metric (similarity) learning framework. Classification and state estimation were integrated into a Siamese network~\cite{SiamFC} that measures the similarity between the target and the candidates for tracking. Semantic branches and appearance branches were constructed in the dual Siamese network~\cite{SASiam}, and saliency mechanisms were introduced in the attention-based Siamese network~\cite{RASNet}. SiamRPN~\cite{SiamRPN,SiamRPN++} combined the Siamese network with the region proposal network (RPN) to allow trackers estimating target extent when positioned accurately. SiamMask~\cite{SiamMask} involved a unified framework for visual target tracking (VOT) and video object segmentation (VOS). To put it simply, a tracker is trained offline, which relies on the position specified by the first frame for semi-supervised learning to achieve target tracking and mask estimation. 
Despite of the efficiency, Siamese trackers are less robust to the interference from background due to ignoring offline training. ATOM~\cite{ATOM} solved this issue by using a large number of samples for offline training. Nevertheless, with multiple sampled features, the target response map could have multiple sub-peaks, which aggregates the risk of model drift, particularly when there is interference from target appearance variation and/or background noise.



\section{Methodology}
We propose the Sub-Peak Response Suppression tracker (SPSTracker), with Peak Response Pooling (PRP) and Boundary Response Truncation (BRT) modules, to aggregate the multiple sub-peaks on a tracking response map into a single enforced peak, as shown in Fig.~\ref{fig:method}. SPSTracker is built upon the ATOM tracker~\cite{ATOM}, with a target classification branch and a target localization branch. The classification branch converts the feature map into a response map and provides the coarse locations of the target. The localization branch uses the bounding-box regression to localize targets. Upon the classification branch, the PRP and BRT modules are applied in a plug-and-play manner, as shown in Fig.~\ref{fig:flowchart}.

\subsection{Tracking Response Prediction}
The classification branch is a CNN online structure, which learns from minimizing the tracking response and Gaussian priori $y_j$. Denote the feature map of a current video frame (the test image) from CNN as $x$. The classification branch is a 2-layer fully convolutional network parameterized with $w$, which predicts the tracking response map $f(x;w)$, as
\begin{equation}
f(x;w)=\phi _{2}(w_{2}\ast \phi _{1}(w_{1}\ast x)), \label{E1}
\end{equation}
where $w_{1}$ and $w_{2}$ denote parameters for first and second convolutional layers, respectively, the symbol of $*$ denotes standard multi-channel convolution, and $\phi _{1}$ and $\phi _{2}$ are the activation functions.

During object tracking, the parameters of the classification branch are updated by minimizing the following objective function:
\begin{equation}
L(w)=\sum_{j=1}^{m}\gamma _{j}\left \| f(x_{j};w)-y_{j} \right \|^2+\sum_{k}\lambda _{k}\left \| w_{k} \right \| ^2 \label{E2},
\label{eq.cls}
\end{equation}
where $j$ denotes the index of training samples,  $x_{j}$ denotes the features from the $j^{th}$ sample, $y_j$ set to a sampled Gaussian prior at the target location~\cite{ATOM},as shown in Fig~\ref{fig:method}(Gaussian prior). 
$\gamma_{j}$ denotes the weight of the corresponding training sample, and $\lambda_{k}$ is a parameter to trade-off the contributions of the two terms.

By optimizing Eq.~(\ref{eq.cls}) with a conjugate gradient descent method, the model predicts the target response map, as shown in Fig.~\ref{fig:method}. Due to the response map is a weighted sum of response maps from multi-scale samples and thereby appears a multi-peak distribution. This makes the maximum response not consistent with the target geometric centers, thus increasing the classification error and the risk of model drift.

\subsection{Sub-Peak Response Suppression}

To conquer the issue that Sub-Peak Response causes the model drift, we propose Sub-Peak Response Suppression method, which can prevent the sub-peak from ``growing" into the main-peak.
 Specifically, we directly operate the target response map predicted by $f(x_j;w)$ and reformulate Eq.~(\ref{E2}) as 
\begin{equation}
L(w)=\sum_{j=1}^{m}\gamma _{j}\left \| f'( x_{j};w)-y_{j} \right \|^2+\sum_{k}\lambda _{k}\left \| w_{k} \right \| ^2,  \label{E3}
\end{equation}
\begin{equation}
f'( x_{j};w) = \beta_{j} [ f ( g ( x_{j} ) ; w )+ P ( f (g ( x_{j}) ; w))], 
 \label{E4}
\end{equation}
where $P$ denotes the Peak Response Pooling applied on each sampled response map, $g( x_j)$ denotes the feature after Boundary Response Truncation (BRT) operation which decreases the variance of target response and reduces the boundary effect, and $\beta _{j}$ denotes the weight for the $j^{th}$ sample. By using PRP and BRT, we can apply feature fusion to aggregate the response maps from multiple samples into the target response map, as well as guaranteeing the the response map has a single peak centered at the target. 
\begin{figure*}[!t]
\begin{center}

{\includegraphics[width=1.4\columnwidth]{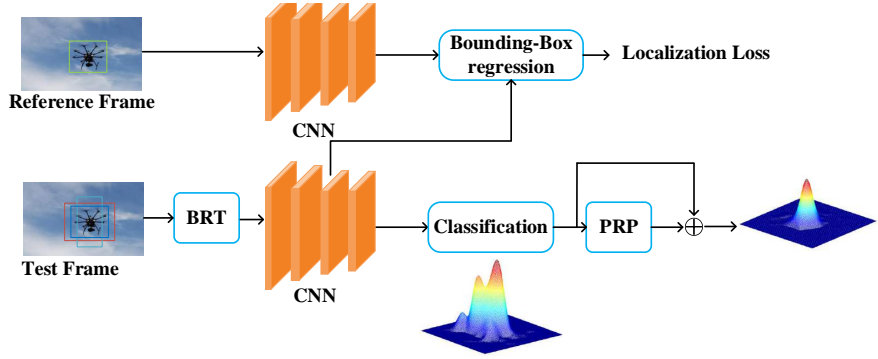}}
\end{center}
   \caption{The flowchart of the proposed SPSTracker. It has a target classification branch and a target localization branch. The classification branch converts the feature map into a response map and provides the coarse locations of the target. The localization branch uses bounding-box regression to localize targets. Upon the classification branch, the PRP and BRT modules are applied in a plug-and-play manner.}
\label{fig:flowchart}
\end{figure*}

By minimizing the objective function of Eq.~(\ref{E2}), we can force the response map $f(x_j;w)$ to approximate the Gaussian priori $y_j$. However, for the targets of partial occlusion or background noises, $f(x_j;w)$ could be acenteric and unlikely to be a Gaussian distribution. The operations of PRP and BRT in Eq.~(\ref{E4}) make $f'(x_j;w)$ close to the Gaussian prior $y_j$ via Eq.~(\ref{E3}), and eventually facilitates the online learning procedure.

\subsubsection{Peak Response Pooling (PRP)}
We propose a Peak Response Pooling (PRP) module, which concentrates the maximum values on the tracking  response map to the target geometric center.
On the target response map output by the classification branch, horizontal PRP is first performed to concentrate the response map into a horizontal pooling map. This procedure is done by finding the maximum response in each row of the response map and assigns all pixels of the line the maximum response value.
In a similar way, vertical PRP is performed in each column on the response map to obtain the vertical pooling map. As a result, the element value of the response map after the PRP operation can be calculated as
\begin{equation}
\widehat{x}_{pq}\!=\!\max(x_{p1},x_{p2},...,x_{pn})+\max(x_{1q},x_{2q},...,x_{mq}) \label{E5}
\end{equation}
where $x_{pq}$ denotes the original response value at the location of the $p^{th}$ row and the $q^{th}$ column.
The horizontal and vertical pooling maps are summed to obtain the rectified response map, which tends to aggregate large response values to the target geometric center.
After multiple learning iterations, the target response is concentrated to approximate a 2D Gaussian distribution, which fits the Gaussian priori distribution for robust object tracking, as shown in Fig.~\ref{fig:method}. 

The Peak Response Pooling (PRP) is inspired by the center/corner pooling~\cite{CenterNet2019,CornerNet2018} developed for object detection. However, PRP is different from the center/corner pooling from the following two aspects: 1) PRP targets at aggregating the response map to a single-peak distribution so that the Gaussian prior distribution can be well fitted. In contrast, the center/corner pooling aligns features to handle the appearance variance of objects; 2) PRP leverages more efficient row- and column-wise maximization operations to aggregate the large response to target centers, while the center/corner pooling uses comparison and substitution operations.


\subsubsection{Boundary Response Truncation (BRT)}
When recognizing objects and determining an objective boundary, the human visual system does not align objects with some fixed data points but uses Fovea in eyeballs that concentrate peak response to central regions for object localization~\cite{kong2019foveabox}. This concentration procedure inspires us to develop the BRT module for object tracking.

During tracking, for the pixel on the extent of the target but far away target centers could have ambiguous features (either background or foreground). The PRP module can concentrate the target response to the target centers but does not consider the variance of the target response. In complex scenes, the response maps could have large variance for the significant response from the target boundary, which is called the boundary effect. Considering that a single-peak response map with small variance could alleviate the boundary effect and improve the tracking robustness, we further introduce the Boundary Response Truncation (BRT) operation.

As shown in Fig.~\ref{fig:method}, BRT is a simple clip operation, which sets the pixels far away from the peak response to be zero. This operation discards the response at the target boundary and reduces the variance of the response map. With BRT, we may miss some informative target response. However, it is experimentally validated that clipping the response map by 10\%; we lose 4\% foreground information and 12\% background information, \textit{i.e.}, BRT reduces more ambiguous response while enhancing the classification ability of the tracker.

\subsection{Object Tracking}

SPSTracker is built upon the state-of-the-art ATOM tracker~\cite{ATOM}, with a target classification branch and a target localization branch. The classification branch produces coarse region proposals by evaluating the target response map. The target localization branch fine-tunes the network parameters to fit the reference target box with multiple region proposals~\cite{ATOM}. Upon the classification branch, the PRP and BRT modules are applied in a plug-and-play manner, as shown in Fig.~\ref{fig:flowchart}.

Note that the ATOM tracker uses only the last convolutional layer of ResNet-18 (Block4) as feature representation. Shallow convolutional features are more important for extracting some low-level information such as color and edge, while deep convolutional features are rich in high-level semantics. The fusion of multi-scale (shallow and deep) features enforces the representation capability, but it produces sub-peaks on response maps and deteriorates the tracking performance.

By introducing the PRP and BRT modules, the multi-scale features can be well integrated for target representation and tracking. As shown in Fig.~\ref{fig:sps}, the multiple sub-peaks produced by multi-scale features can be concentrated into a maximum peak, which bridges the gap between $f'(x_j;w)$ and $y_j$ and facilitates robust tracking.

\begin{table}[t]
\begin{center}
\resizebox{\columnwidth}{!}{
\begin{tabular}{ccc|cccc}
\hline
BRT & PRP & MF & EAO & Accuracy & Robutness & FPS\\
\hline
& & & 0.401& 0.590 & 0.204 &\textbf{40} \\
\checkmark &  & &0.414& 0.601 & 0.189 &\textbf{40} \\
  & \checkmark & & 0.420&  0.609 & 0.191 &39 \\
& & \checkmark & 0.411 & 0.605& 0.184 &35\\
\checkmark & \checkmark & &  0.424& 0.610 & 0.179 &39\\
\checkmark & & \checkmark  &0.419 & 0.604  & 0.187 &35\\
& \checkmark & \checkmark &  0.424 & 0.612  & 0.174 &35\\
\checkmark& \checkmark& \checkmark  & \textbf{0.434}& \textbf{0.612} & \textbf{0.169} &35\\
\hline
\end{tabular}
}
\end{center}
\caption{Ablation study of SPSTracker on the VOT-2018 benchmark. ``BRT'' denotes Boundary Reponse Truncation, ``PRP'' denotes Peak Response Pooling, and ``MF'' multi-scale feature fusion. The baseline performance is reported by the state-of-the-art ATOM tracker.}
\label{tab:ablation}
\end{table}

\section{Experiment}
In this section, we first describe the implementation details of SPSTracker. We then present the ablation study to validate the PRP and BRT modules proposed in this paper. At last, we evaluate the SPSTracker on commonly used benchmarks and compare it with state-of-the-art trackers.
All the experiments are carried out with Pytorch on a Intel i5-8600k 3.4GHz CPU and a single Nvidia GTX 1080ti GPU with 24GB memory.

\subsection{Implementation details}

SPSTracker is implemented upon the ATOM architecture~\cite{ATOM}, by using ResNet-18~\cite {ResNet} pre-trained on ImageNet as the backbone network. 
The Block3 and Block4 features extracted from the test image are first passed through two Conv layers. Regions defined by the input bounding boxes are then pooled to a fixed size using pooling layers. The pooled features are modulated by channel-wise multiplication with the coefficient vector returned by the reference branch. The features are then passed through fully-connected layers to predict the Intersection over Union (IoU). All Conv and FC layers are followed by BatchNorm and ReLU. The target response map is obtained by fusing the response obtained by ResNet's block3 and block4. 
\begin{figure}[t]
\includegraphics[width=.95\columnwidth]{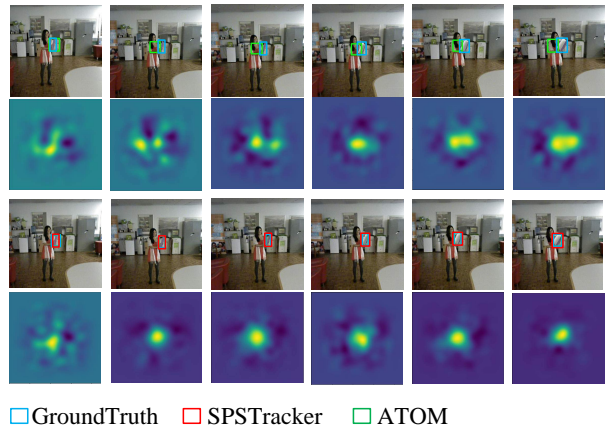}
\caption{Comparison of the target response maps of the ATOM tracker (up) and SPSTracker (down). }
\label{fig:sps}
\end{figure}
\subsection{Ablation Study}

For the proposed PRP and BRT modules, we perform ablation analysis to investigate their impact on the tracking performance. We also analyze the impact of the multi-scale features in SPSTracker. All the ablation studies are carried out on the VOT2018~\cite{vot2018} benchmark.

\textbf{Peak Response Pooling (PRP).} From the results in Table ~\ref{tab:ablation}, we can see that the introduction of PRP module to the classification branch significantly aggregates the tracking performance. Specifically, it improves the expected average overlap (EAO) value by 0.19 (0.401 to 0.420), which is a significant margin, considering the strong baseline ATOM. It also improves the tracking accuracy and robustness as indicated by the last two rows of Table ~\ref{tab:ablation}.

\begin{figure} [t]
\centering    
\subfloat{
 \label{fig:a}     
\includegraphics[width=0.45\columnwidth,height=0.40\columnwidth]{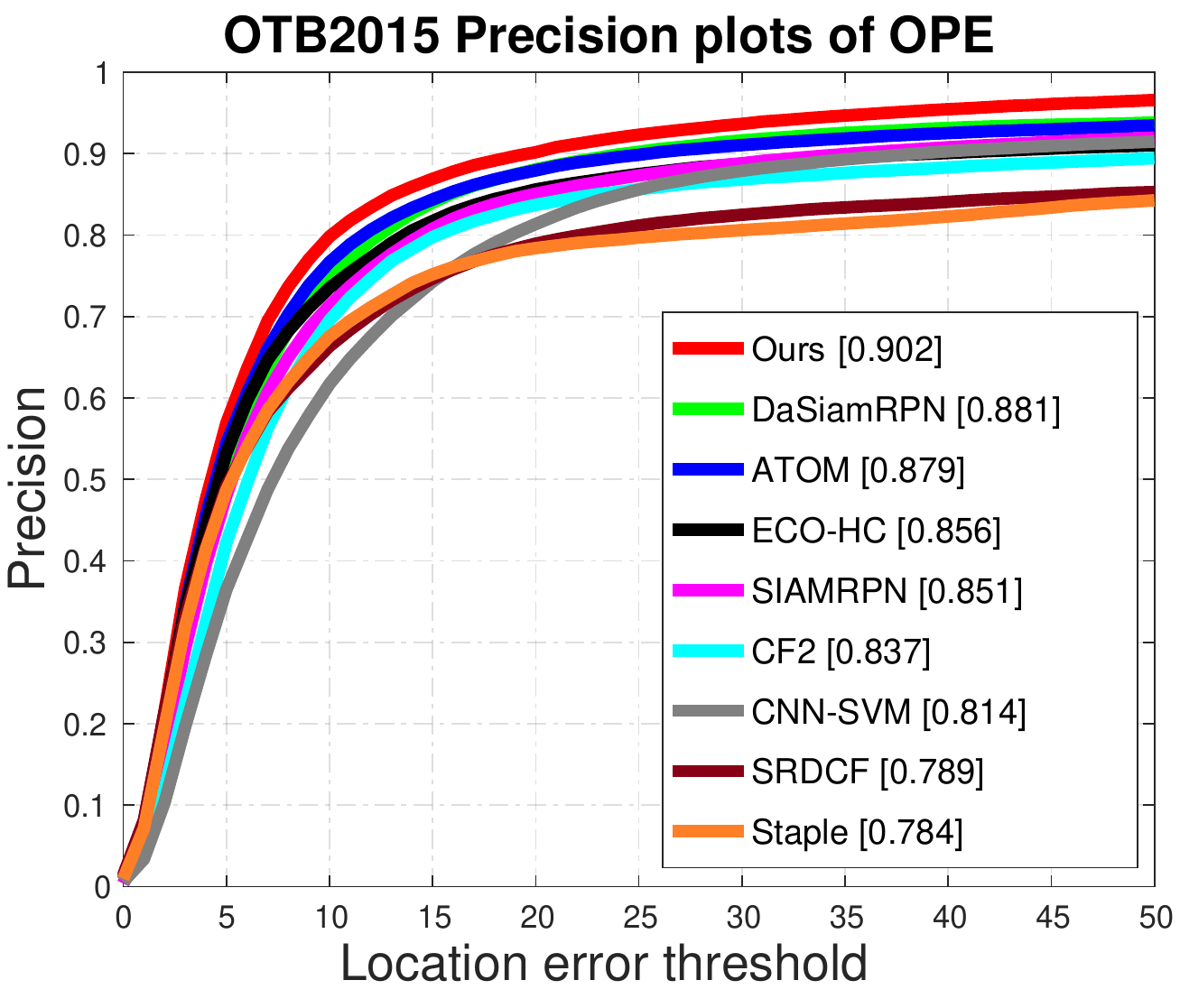}  
}
\subfloat{ 
\label{fig:b}     
\includegraphics[width=0.45\columnwidth,height=0.40\columnwidth]{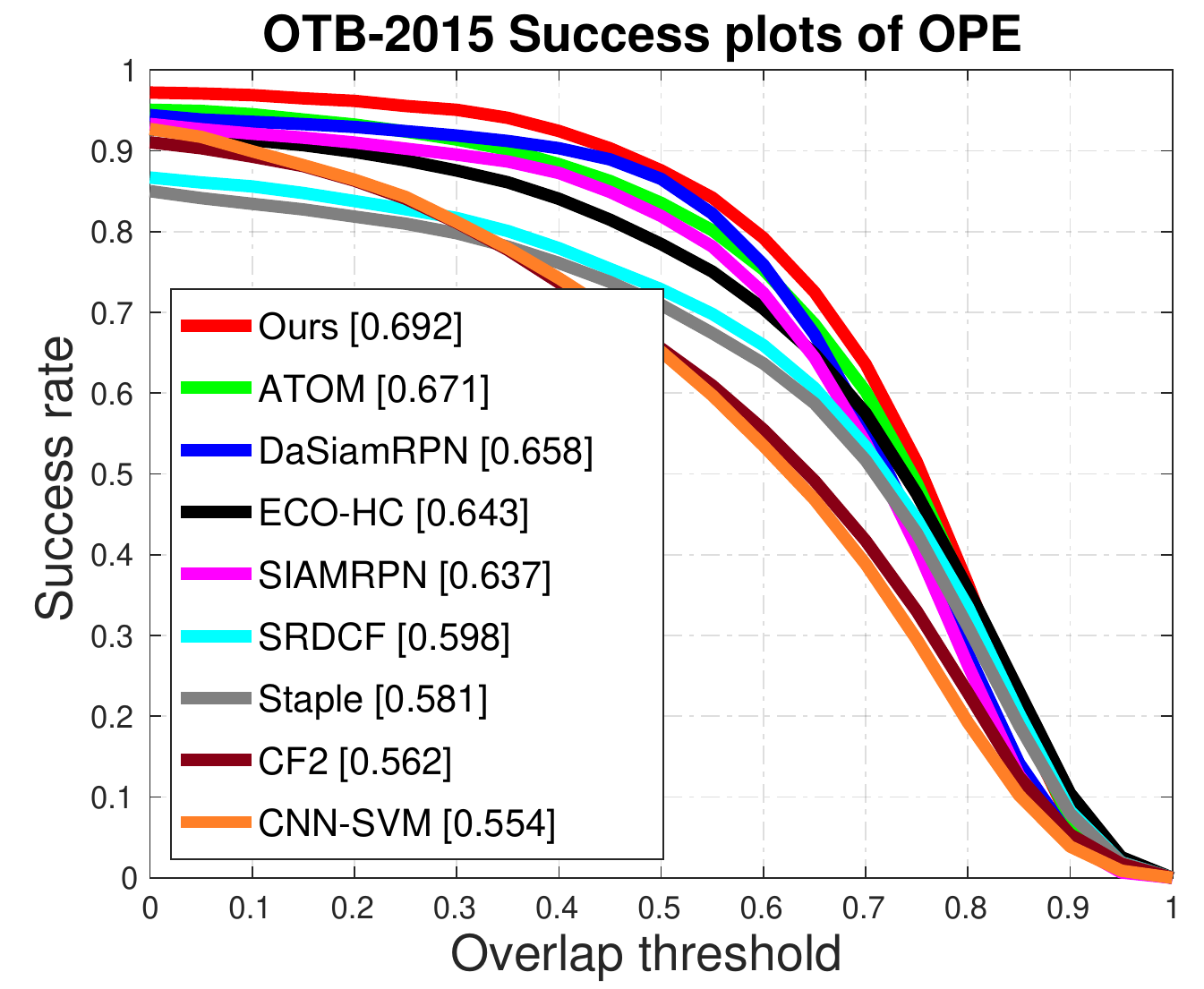}     
} 
\quad
\subfloat{ 
\label{fig:c}     
\includegraphics[width=0.45\columnwidth,height=0.40\columnwidth]{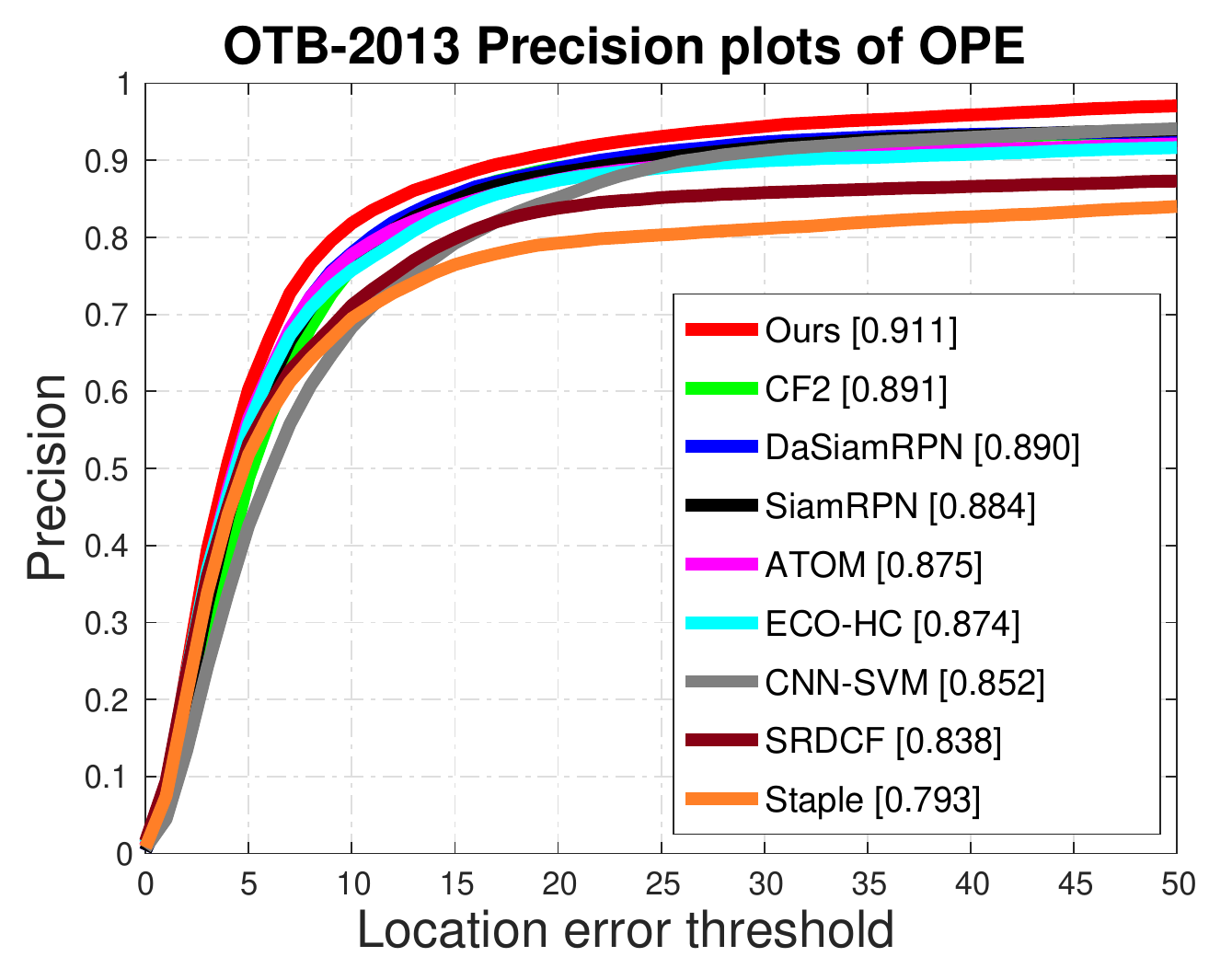}     
}   
\subfloat{ 
\label{fig:d}     
\includegraphics[width=0.45\columnwidth,height=0.40\columnwidth]{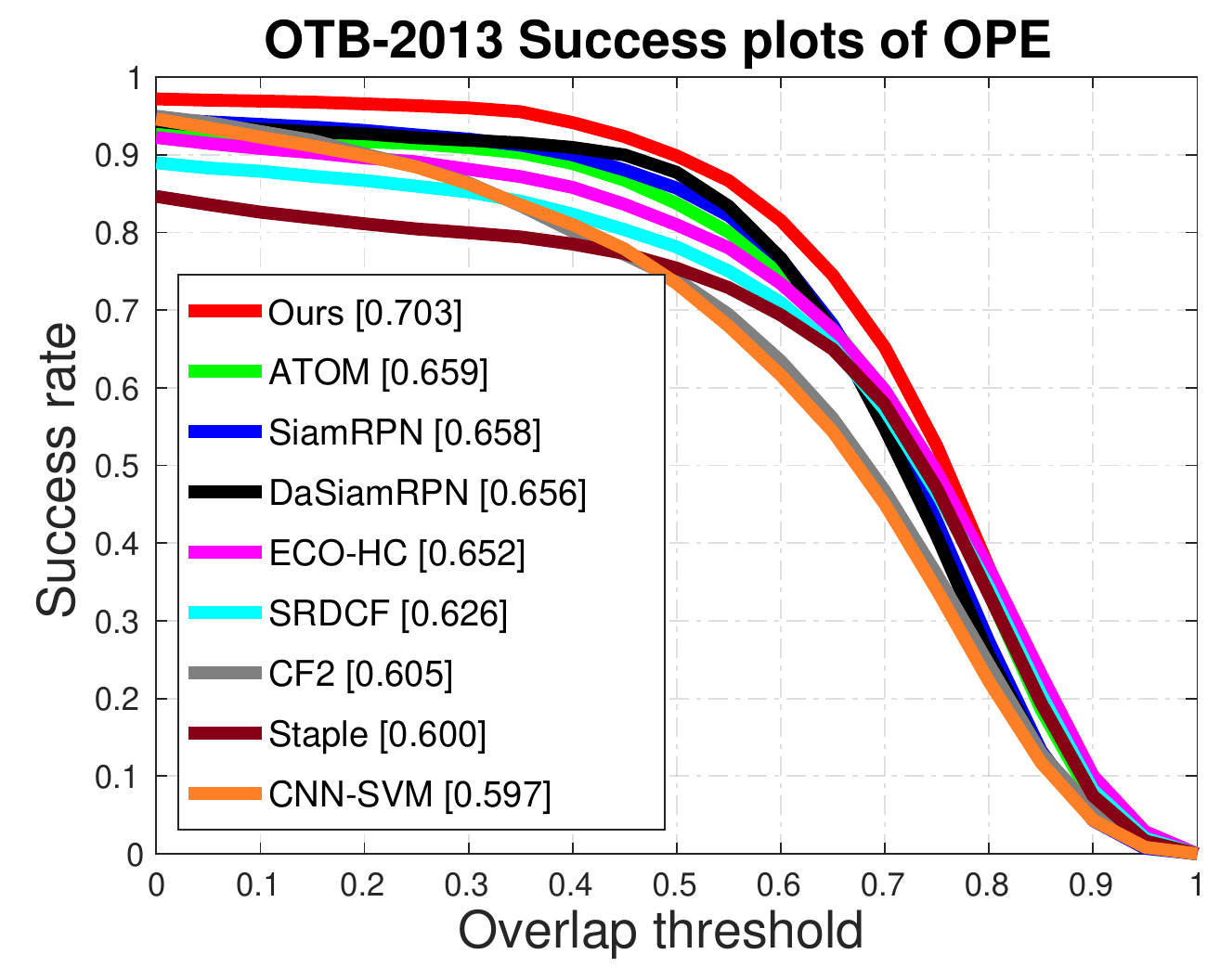}     
} 
\quad
\subfloat{ 
\label{fig:e}     
\includegraphics[width=0.45\columnwidth,height=0.40\columnwidth]{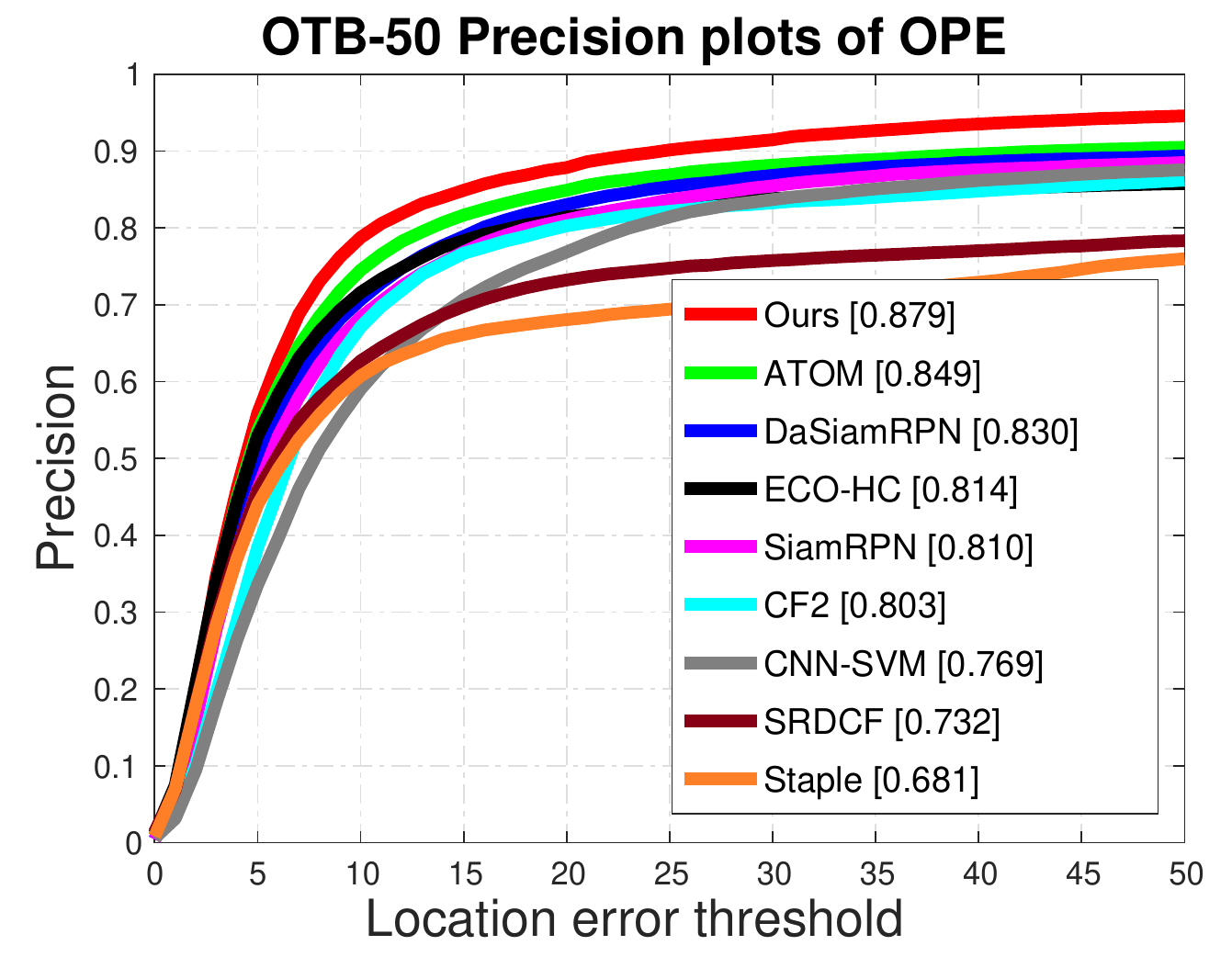}     
}   
\subfloat{ 
\label{fig:f}     
\includegraphics[width=0.45\columnwidth,height=0.40\columnwidth]{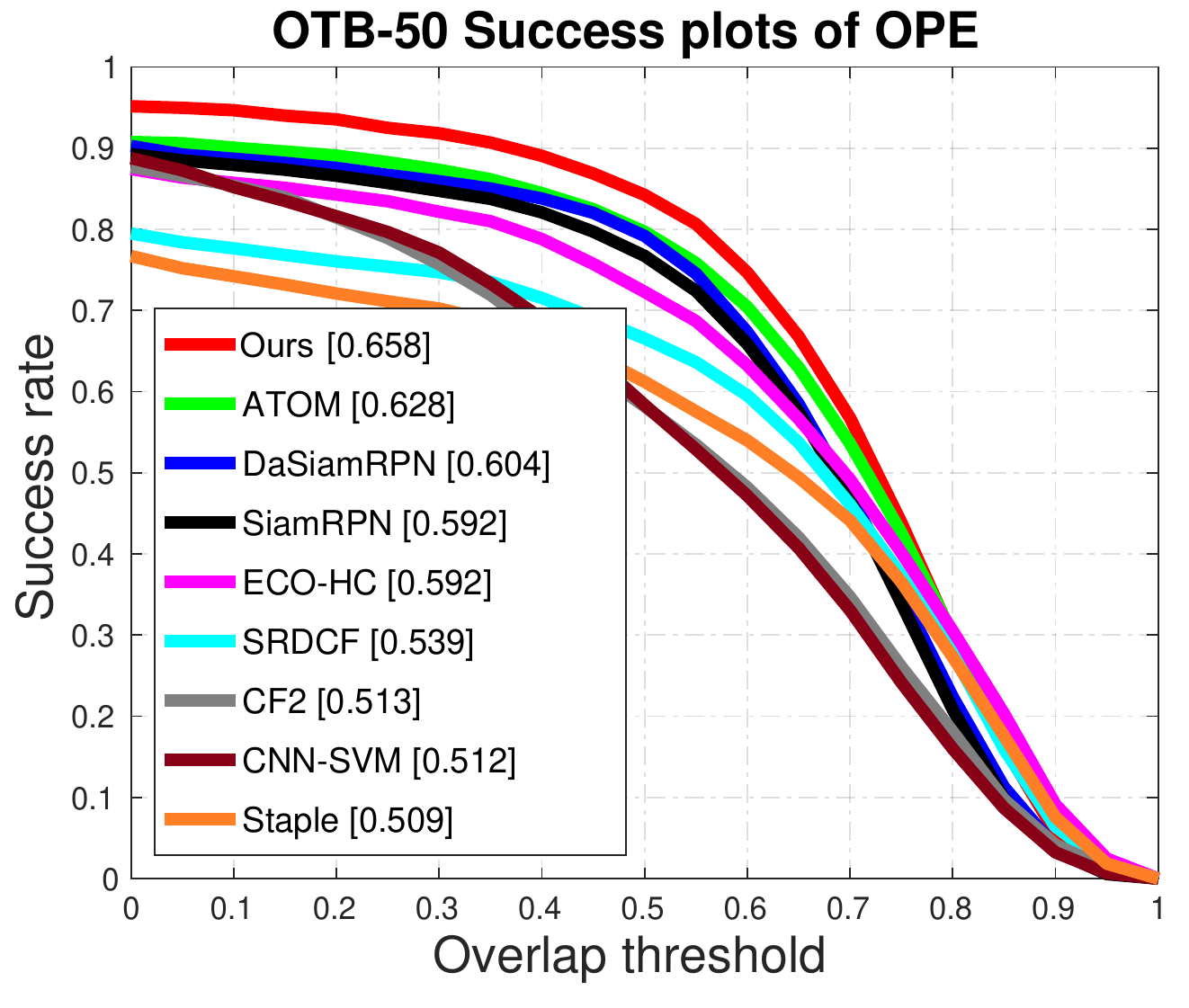}     
}
\caption{ The precision plots and success plots on OTB-2015, OTB-2013 and OTB-50 benchmarks.}     
\label{fig:otb}     
\end{figure}

\begin{figure}[!ht]
\begin{center}
   \includegraphics[width=.95\columnwidth]{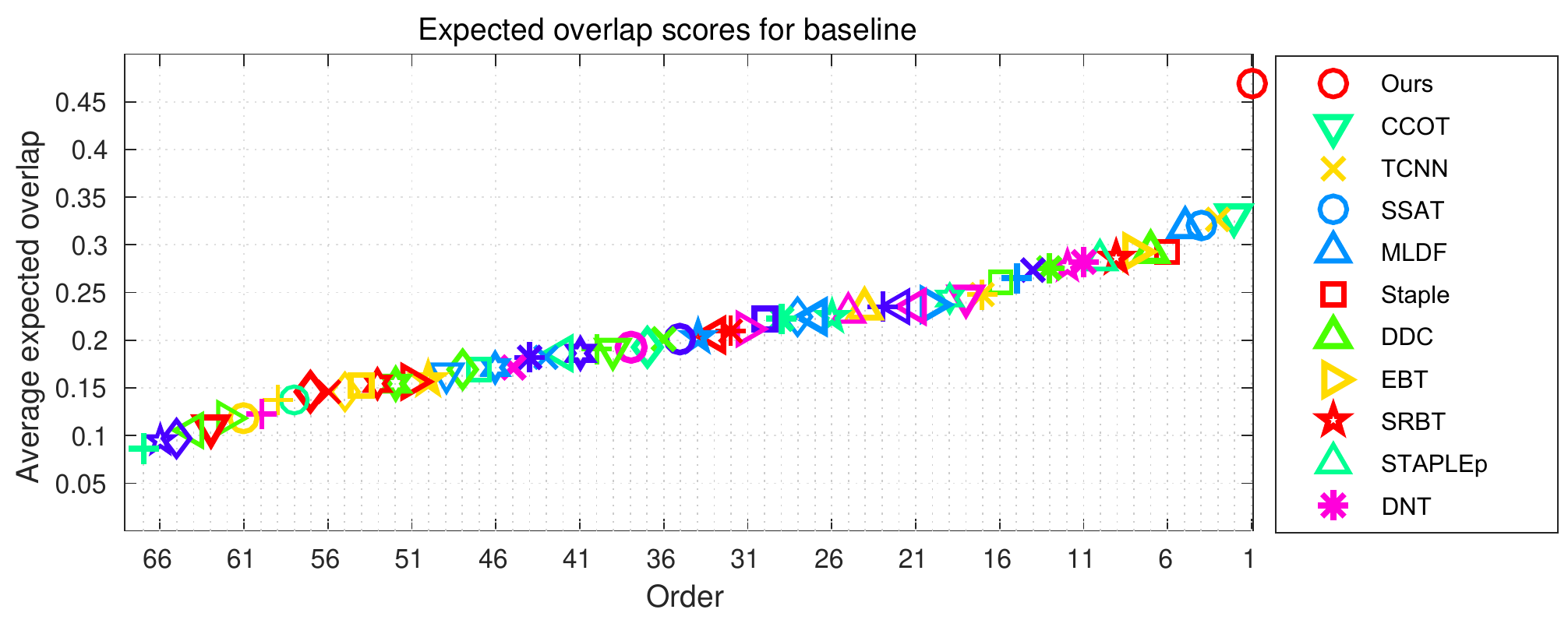}
\end{center}
   \caption{EAO ranking of the tested trackers on VOT2016. 
}
\label{fig:vot2016}
\end{figure}

\begin{figure}[!ht]
\begin{center}
   \includegraphics[width=.95\columnwidth]{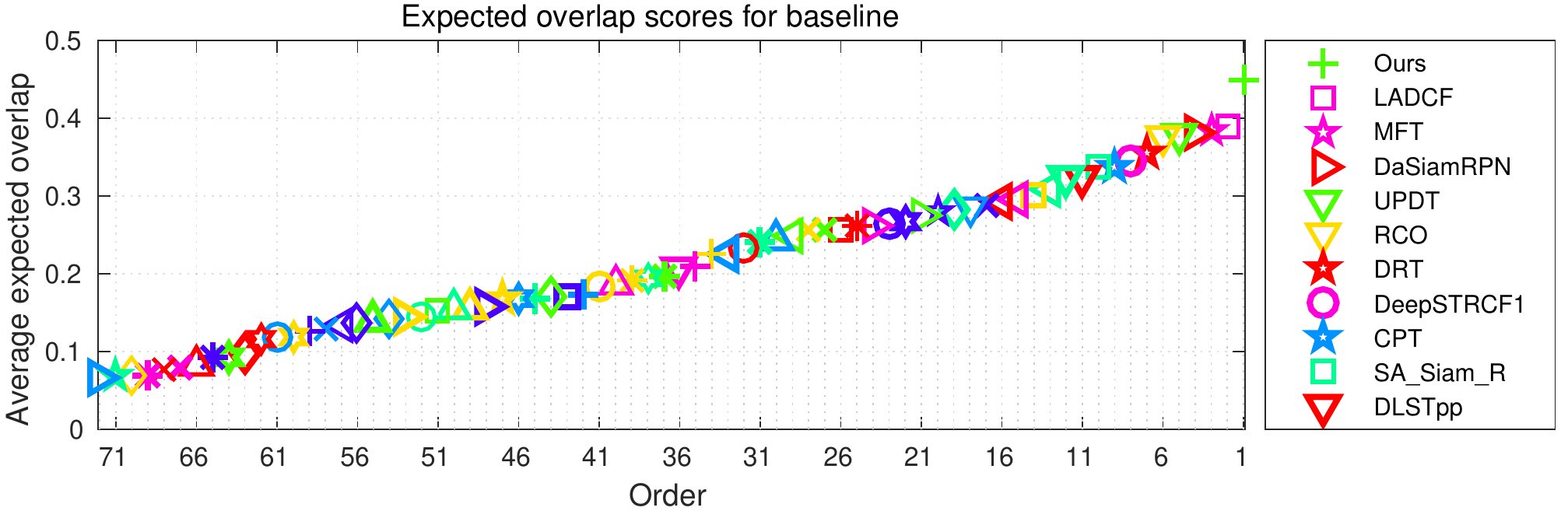}
\end{center}
   \caption{EAO ranking of the tested trackers on VOT2018. 
   }
\label{fig:vot2018}
\end{figure}
\textbf{Boundary Response Truncation (BRT).} The BRT module improves the EAO value by $0.13$ ($0.401$ to $0.414$), as reported in Table ~\ref{tab:ablation}, which is also a significant margin. This validates that the truncation operation is able to eliminate response variance and benefit online classifier learning by filtering out ambiguous samples.

We test the truncation size and validate that the best performance is obtained when clipping $10\% \sim 12.5\%$ width/height of the response map. For all the experiments, we clip a $10\%$ height/width of the response map.

\textbf{Multi-scale Feature Fusion.} By using the multi-scale feature fusion, we improve the EAO value by $0.1$. Combining feature fusion with PRP and BRT modules, we can improve EAO by $0.33$ ($0.434$ vs. $0.401$), as reported in Table ~\ref{tab:ablation}. The significant performance gain demonstrates that the proposed PRP and BRT modules facilitate the fusion of multi-scale features, and reduce the negative effect brought by the multiple sub-peaks and the feature fusion.

\begin{table}[!t]
\begin{center}
\begin{tabular}{l|ccc}
\hline
Tracker & EAO & Accuracy & Robustness \\
\hline\hline
SPSTracker  &{\color{red}{0.459}}  & {\color{blue}{0.625}}  & {\color{red}{0.158}}  \\
\hline
SiamMask & {\color{blue}{0.433}} & {\color{red}{0.639}} &  {\color{blue}{0.214}}\\
DWSiam & {\color{green}{0.370}} & {\color{green}{0.580}} & 0.240\\
CCOT & 0.331 & 0.539 &0.238 \\
TCNN & 0.325 & 0.554 & 0.268 \\
SSAT & 0.321 & 0.577 & 0.291\\
MLDF & 0.311 & 0.490 & {\color{green}{0.233}}\\
Staple & 0.295 & 0.544 & 0.378\\
DDC & 0.293 & 0.541 & 0.345\\
EBT & 0.291 & 0.465 & 0.252 \\
SRBT & 0.290 & 0.496 & 0.350\\
\hline
\end{tabular}
\end{center}
\caption{Performance comparison on VOT2016.}
\label{vot2016}
\end{table}

\begin{table}[!t]
\begin{center}
\begin{tabular}{l|ccc}
\hline
Tracker & EAO & Accuracy & Robustness \\
\hline\hline
SPSTracker  &{\color{red}{0.434}}& {\color{red}{0.612}}& 0.169  \\
\hline
SiamRPN++ & {\color{blue}{0.414}} & {\color{green}{0.600}} & 0.234\\
ATOM & {\color{green}{0.401}}&  0.590 & 0.204 \\
SiamMask & 0.380 & {\color{blue}{0.609}}& 0.276 \\
LADCF &  0.389 & 0.503 & {\color{green}{0.159}}\\
MFT & 0.385 & 0.505  & {\color{red}{0.140}}\\
DaSiamRPN & 0.383& 0.544 & 0.276\\
UPDT &  0.378 & 0.536 & 0.184\\
RCO &0.376  &  0.507  & {\color{blue}{0.155}}\\
DRT &0.356 & 0.519 & 0.201\\
DeepSTRC &  0.345 &0.523  & 0.215\\
\hline
\end{tabular}
\end{center}
\caption{Performance comparison on VOT-2018. }
\label{tab:vot2018}
\end{table}

\textbf{Sub-peak suppression.}
In Fig.~\ref{fig:sps}, we compare the target response maps of the ATOM tracker and SPSTracker. It can be seen that SPSTracker suppresses multiple sub-peaks while producing the response map of a single peak centered at the target. The peak response can well fit a Gaussian distribution prior $y_j$.

\textbf{Tracking speed.}
With a single GPU, the proposed SPSTracker achieves a tracking speed at $35$ fps. Compared with the speed ($40$ fps) of the baseline ATOM, SPSTracker achieves significant performance gains with a negligible computational cost overhead.


\textbf{OTB.} The object tracking benchmarks (OTB)~\cite{otb2013,otb2015} consists of the three datasets, namely OTB-2013~\cite{otb2013}, OTB-50 and OTB-100 which consist of 51, 50 and 100 fully annotated videos, respectively. OTB100 includes OTB2013 and OTB50. All sequences belong to 11 typical tracking interference properties.
\begin{table*}[ht]
\begin{center}
\begin{tabular}{l|ccccccccccc}
\hline
 &Ours& ATOM & UPDT & CCOT &ECO & MDNet & HDT &DaSiamRPN & FCNT & SRDCF  &BACF \\
\hline
AUC&60.0 &59.0&54.2&49.2&47.0&42.5&40.0&39.5&39.3&35.3&34.2\\
\hline
\end{tabular}
\end{center}
\caption{Performance comparison on the NFS dataset.}
\label{tal:nfs}
\end{table*}

\begin{figure*}[ht]
\begin{center}
  \centerline{\includegraphics[width=1.8\columnwidth]{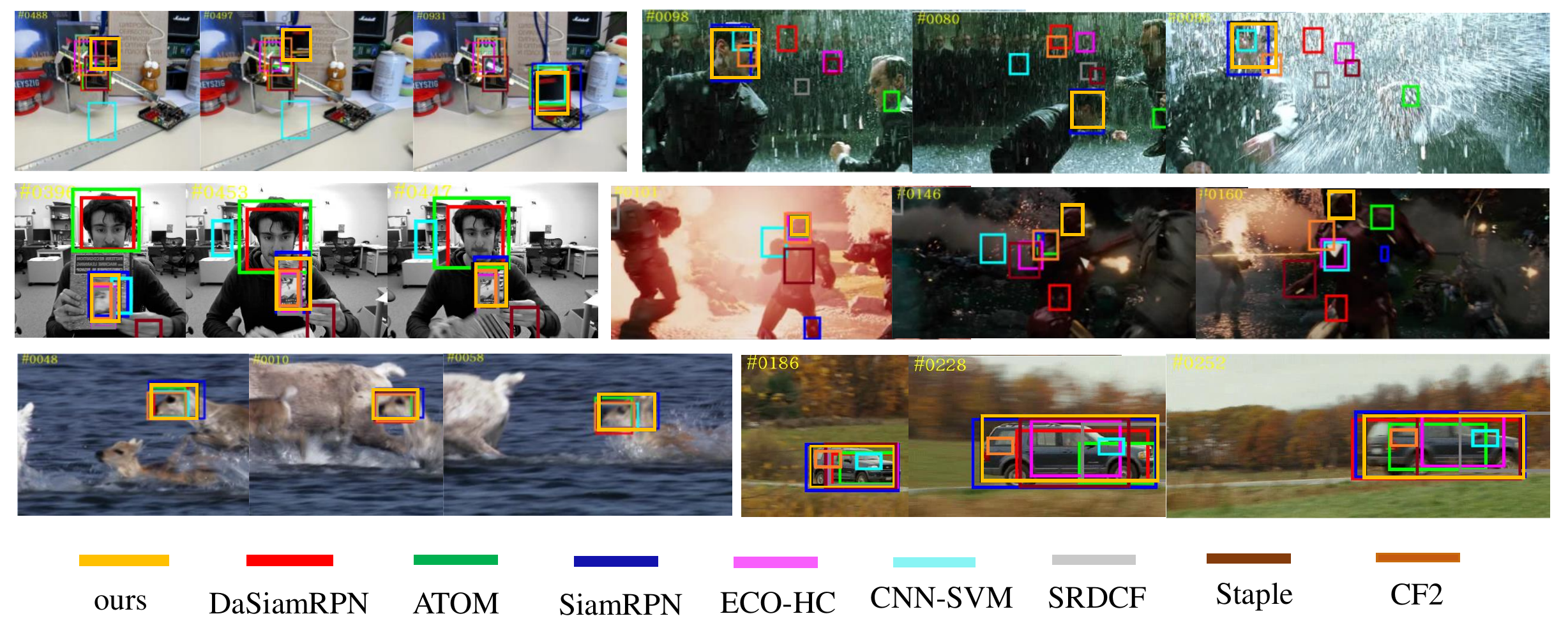}}
\end{center}
\caption{Qualitative results of state-of-the-art trackers on sequences \emph{Box, Matrix, ClifBar, Ironman, Deer and CarScale}. SPSTracker can localize objects with interference from either foreground or backgrounds. In contrast, other compared methods have failure cases. (Best viewed in color with zoom in)}
\label{fig:res}
\end{figure*}

Two evaluation metrics, success rate and precision, are used on OTB. The precision plot shows the percentage of frames whose tracking results are within a certain distance, which is determined by a given threshold. The success plot shows the ratio of successful frames when the threshold changes from 0 to 1, where a successful frame indicates that its overlap is greater than the given threshold. The area under the curve (AUC) of each success plot is used to rank the tracking methods.

By using the success rate and precision plot in the one-pass evaluation (OPE) as the evaluation metric, we compare the SPSTracker with state-of-the-art trackers including ATOM~\cite{ATOM}, DaSiamRPN~\cite{DaSiamRPN}, ECO-HC~\cite{ECO}, SiamRPN~\cite{SiamRPN}, CF2~\cite{CF2}, CNN-SVM~\cite{CNN-SVM}, SRDCF~\cite{SRDCF} and Staple~\cite{Staple}. As shown in Fig.~\ref{fig:otb}, the proposed SPSTracker achieves the best performance on the three benchmarks, by obtaining $0.692$, $0.703$ and $0.658$ AUC scores on OTB-2015 and OTB-2013, OTB-50, respectively. Compared with ATOM~\cite{ATOM}, SPSTracker improves by $2.1\%$, $4.4\%$ and $3.0\%$, respectively.

\textbf{VOT2016 and VOT2018.}
From the visual object tracking (VOT) benchmark, we select VOT2016~\cite{vot2016} and VOT2018~\cite{vot2018} to evaluate the trackers. VOT2016 contains 60 challenging videos, while VOT2018 includes 10 more challenging sequences.  Whenever the tracking bounding box drifts way from the ground truth, the tracker re-initializes after five frames. The trackers are evaluated by the EAO metric, which is the inner product of empirically estimated average overlap and typical sequence length distribution. In addition, accuracy (average overlap) and failures/robustness (average number of failures) are used for evaluation as well. 

SPSTracker is compared with 10 state-of-the-art trackers on VOT-2016, as shown in Fig.~\ref{fig:vot2016}. SPSTrack achieves the leading performance and significantly outperforms other trackers. Table~\ref{vot2016} reports the details of the comparison with  SiamMask~\cite{SiamMask},DWSiam~\cite{DWSiam}, CCOT~\cite{CCOT}, TCNN~\cite{Nam2016}, SSAT~\cite{vot2016},MLDF~\cite{vot2016}, Staple~\cite{Staple}, DDC~\cite{vot2016}, EBT~\cite{EBT} and SRBT~\cite{vot2016}. 
The EAO score of the proposed SPSTracker is $0.459$, which is significantly higher than the peer trackers.

SPSTracker is also compared with the 10 state-of-the-art trackers on VOT-2018. As shown in Fig.~\ref{fig:vot2018}, SPSTracker also obtains the best performance. Table~\ref{tab:vot2018} shows the details of the comparison.

SPSTracker achieves an EAO score of $0.434$, which is significantly better than that of SiamRPN++~\cite{SiamRPN++}, ATOM and other state-of-the-art trackers. Particularly, it outperforms the state-of-the-art SiamRPN++ by $2$ , ATOM  by $3.3$ and SiamMask by $5.4$, which are significant margins for object tracking on the challenging benchmark.

\textbf{NFS.}
The Need for Speed (NFS)~\cite{NFS} dataset consists of 100 videos (380K frames).
All frames are annotated with axis-aligned bounding boxes, and all sequences are manually labeled with nine visual attributes, including occlusion, fast motion, background clutter. We evaluate the trackers on the 35 FPS version of the NFS dataset. Table~\ref{tal:nfs} reports the AUC scores of the compared trackers. SPSTracker slightly outperforms the baseline ATOM tracker, while significantly outperforms other state-of-the-art tracking methods.

Fig.~\ref{fig:res} shows tracking examples on the OTB benchmark, from which we can see that SPSTracker correctly localizes the targets under serious interference from foreground and backgrounds. In contrast, other trackers have failure cases. 

\section{Conclusions}

Visual tracking has been extensively investigated in the past few years. Nevertheless, the problem about how to model interference from multiple targets, appearance variance and/or background noise remains unsolved. In this paper, we proposed  modeling the interference from the perspective of peak distribution and designed a rectified online learning approach for sub-peak response suppression and peak response enforcement. We proposed plug-and-play Peak Response Pooling (PRP) to aggregate and align discriminative features, and designed Boundary Response Truncation (BRT) to reduce the variance of feature response. Based on PRP and BRT, we integrate multi-scale features in SPSTracker to learn the discriminative features for robust object tracking. SPSTracker achieved the new state-of-the-art performance on six widely-used benchmarks, which verifies the effectiveness of the proposed peak response modeling approach.

{\small
\bibliographystyle{aaai}
\bibliography{1495.references}
}

\end{document}